\documentclass[a4paper]{article}

\usepackage{INTERSPEECH2018}
\usepackage{multirow}
\usepackage{graphicx}
\usepackage{booktabs}
\usepackage{comment}

\title{Joint Learning of Interactive Spoken Content Retrieval \\ and Trainable User Simulator}
\name{Pei-Hung Chung$^1$$^*$\thanks{{$^*$}The two first authors made equal contributions.}, Kuan Tung$^2$$^*$\footnotemark[1], Ching-Lun Tai$^2$, Hung-Yi Lee$^1$\thanks{This work was supported by the Ministry of Science and Technology of Taiwan}}
\address{
  $^1$Graduate Institute of Communication Engineering, National Taiwan University\\
  $^2$Department of Electrical Engineering, National Taiwan University}
\email{\{chung95191, dinotuku, vincent001217, tlkagkb93901106\}@gmail.com}

\begin{document}
\maketitle
\begin{abstract}
User-machine interaction is crucial for information retrieval, especially for spoken content retrieval, because spoken content is difficult to browse, and speech recognition has a high degree of uncertainty.
In interactive retrieval, the machine takes different actions to interact with the user to obtain better retrieval results; here it is critical to select the most efficient action. 
In previous work, deep Q-learning techniques were proposed to train an interactive retrieval system but rely on a hand-crafted user simulator; building a reliable user simulator is difficult. 
In this paper, we further improve the interactive spoken content retrieval  framework by proposing a learnable user simulator which is jointly trained with interactive retrieval system, making the hand-crafted user simulator unnecessary. The experimental results show that the learned simulated users not only achieve larger rewards than the hand-crafted ones but act more like real users. 
%User-machine interaction is crucial for information retrieval, especially for spoken content retrieval. For text retrieval, the retrieved results are readable, so the user can easily scan through and select on a list of retrieved items. This is inaccessible for spoken content or multimedia owing to the difficulty of showing the retrieved item on the screen. Besides, due to the high degree of uncertainty for speech recognition, the first-pass retrieved results can be very noisy. One way to overcome such difficulties is through user-machine interaction. The machine takes different actions to interact with the user to obtain better retrieval results. The suitable actions depend on the retrieval system status, for example requesting for extra information from the user, returning a list of topics for user to select, etc. In our previous work, Deep-Q-Learning techniques are proposed to determine the machine actions. In an interactive spoken content retrieval system, it is impossible for machine to interact with real users and the performance of the retrieval system is affected by both user and agent. Hence, building a reliable user simulator is as challenging as building a good dialog manager. In this paper, we futher modify the framework of the spoken content retrieval system and propose a learnable user simulator which can be jointly trained with a dialog manager. Our experiment results show that the proposed method not only leads to significant performance compared with the previous user simulator and dialogur manager but act more similar to human.  
\end{abstract}

\section{Introduction}
Interactive spoken content retrieval (SCR) system enhances a retrieval system by incorporating user-system interaction~\cite{robins2000interactive,ruthven2008interactive} into the spoken content retrieval process. 
The primary task of SCR is to retrieve the spoken content~\cite{garofolo2000trec} or multimedia~\cite{patton2002interactive,cho2004emotional} desired by the user. 
%After the transcriptions are predicted by ASR, the retrieval system search through the transcribed spoken content corresponding to the query entered by users. Different from text, the retrieval process of spoken content is time consuming and challenging. 
%As a result, for SCR, much more higher retrieval quality is prerequisite and user-system interaction become a promising strategy.
The necessity of user-system interaction in an SCR system is shown in many aspects.
In SCR, the spoken content is usually transcribed into one-best transcriptions or lattices~\cite{saraclar2004lattice,hakkani2006beyond,7114229,can2011lattice,parlak2008spoken}. %so spoken content retrieval greatly depends on the performance of Automatic Speech Recognition (ASR)\cite{7114229,can2011lattice,parlak2008spoken}. 
As ASR errors are inevitable, the retrieved results often contain ambiguities. 
%Additionally, the widely used SCR technologies such like latticed-based indexing result in higher recall rates, but lead to lower precision rates. 
Additionally, it is difficult to display the retrieved multimedia or spoken information items to users~\cite{Zhang:2015:IRC:2766462.2767761}. 
%The importance for user-system interaction is even more obvious on the wearable devices or Internet of things (IoT) devices. This devices usually have no display (or come with only tiny screens), so the retrieval results cannot be displayed as a list for the user to choose from. In addition, multimedia content is usually stored in the cloud instead of local devices. To access the content, wearable or IoT devices must connect to cloud servers through the wireless network. IIR reduces the need to browse multimedia content to search through the desired items, and therefore reduces bandwidth demands. 
%In a nutshell, interactive information retrieval (IIR) boosts the performance and user experience of spoken content retrieval. 

In previous work, the Markov decision process (MDP) has been used to model spoken-content interactive information retrieval (IIR)~\cite{6018284,wen2012interactive,wen2013interactive}. 
Wu et al.~\cite{wu2016interactive} propose using deep reinforcement learning (RL) in interactive SCR. % and have achieved a great improvement. 
Approaches to learning interaction policy include training a system to learn against a user simulator~\cite{schatzmann2006survey,asri2016sequence,li2016user}. 
In spoken-content IIR, the agent learns through interaction; however, it is not feasible to learn this using real users in user-system interaction. Thus a user simulator is used to train spoken-content IIR.
The construction of the user simulator is often the key to a satisfactory system.
	Policies learned with a poor user simulator can be worse than heuristic hand-crafted policies~\cite{thomson2010bayesian}. 
%引用以下這篇 -- Lee
%B. Thomson and S. Young (2010). "Bayesian update of dialogue state: A POMDP framework for spoken dialogue systems." Computer Speech and Language,24(4): 562-588.
In previous approaches, simulated users choose actions based on hand-crafted rules without randomness~\cite{6018284,wen2012interactive,wen2013interactive,wu2016interactive}. 
These hand-crafted rules are not sufficient to simulate the behavior of real users. 
Moreover, simulated users are unrealistic in that they always generate the same responses given the same situations. %in different situations inquire the same query will take exactly the same action, and it makes inconsistency in reality.   

To create a reliable user simulator, we propose a new framework for interactive SCR. 
We design a trainable user simulator for use in joint learning with the dialogue manager, making hand-crafted simulated users unnecessary.
Joint optimization of the dialog agent and the user simulator with deep RL using simulated dialog between the two agents has been proposed for task-oriented dialogue~\cite{liu2017iterative}, and leads to promising improvements after pre-training by supervised learning. 
Learning two agents jointly has also been considered in object detection dialogue~\cite{de2017guesswhat} and negotiation dialogue~\cite{lewis2017deal}.
%interactive SCR has no labeled data for pre-training and the state and action of interactive SCR is limited. 
%It a challenge for a trainable user simulation to achieve significant improvement using finite action (4 in this task) instead of infinite state and action (natural language).
%此處應該強調和前人的不同 -- Lee
%還應該引用 FB 談判那篇、還應該引用 猜東西 那篇 (盧博報過) -- Lee\textbf{d}
In this paper, we consider interactive SCR, which is quite different from previous work.
Here both the user simulator and the IIR system are learned jointly by  deep Q-network (DQN)~\cite{mnih2013playing} from scratch without any labeled data for pre-training.
%The extension to DQN has recently achieve great renown and success on playing Atari games\cite{hessel2017rainbow}. 
We use advanced DQN algorithms~\cite{hessel2017rainbow} to improve the performance of both the interactive SCR system and the user simulator.
The learned simulator displays more sophisticated behavior and acts more like a real user.
%To verify the feasibility of the advanced DQN algorithm, we also modify the retrieval module and attempt other RL-based algorithms to improve the performance of both dialogue manager and user simulator. 
%In this paper, our contribution is twofold. 
%Firstly, we propose a jointly trained user simulator, and the simulator act is more similar to human behavior. Secondly, the extension to DQN is useful for both dialogue manager and user simulator.

%如果文章太長就不需要這段 -- Lee
%The remainder of the paper is organized as shown below. In Section 2, we discuss the proposed framework and DQN setting in detail. In Section 3, we discuss the experiment setup and analyze the results. Section 4 gives the conclusion.

\begin{figure*}[t]
  \centering
  \includegraphics[width=0.97\textwidth]{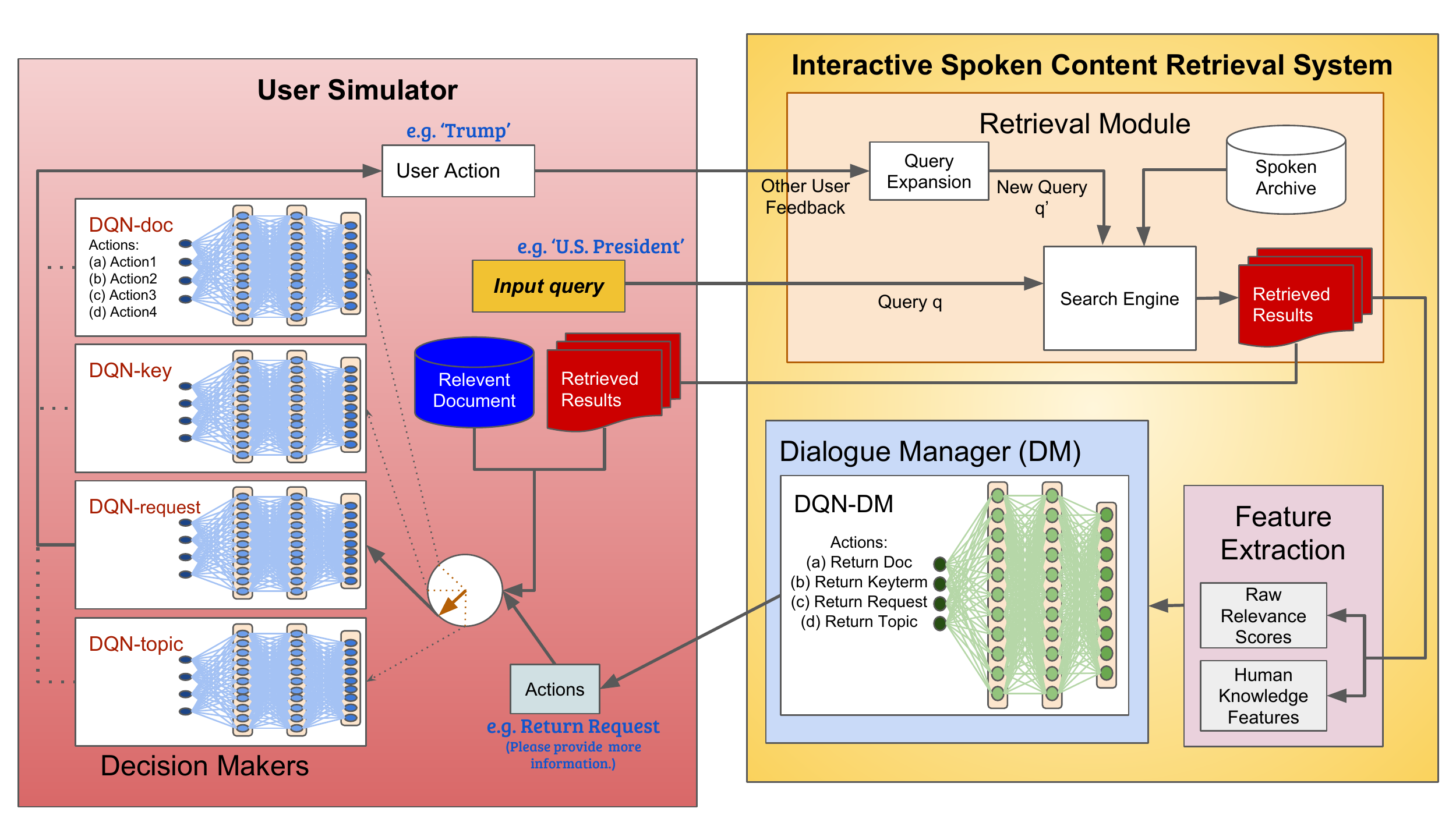}
  \caption{Block diagram of jointly learned user simulator and interactive spoken content retrieval system}
  \label{fig:Proposed ISCR}
\end{figure*}

\section{Proposed Approach}
%不需要比較這篇文章和 吳彥諶 的差別，因為 reader 根本不可能記得吳的 paper 內容 -- Lee

%=============== 我加了這段,讓讀者知道這整個 framework 是什麼 (不要忘了讀者是一次看到這篇文章) ===============
Figure~\ref{fig:Proposed ISCR} illustrates the framework of the proposed interactive SCR, in which the interactive SCR system and the user simulator are jointly learned.
In the proposed framework, the user simulator is given a set of queries and their corresponding relevant spoken documents.
Although the system developers must provide data to the user simulator, they do not need to recruit real users to interact with the system, which greatly facilitates system development.
During training, the user simulator enters a query (e.g., ``US president'') into the SCR system. 
The system retrieves the search results, and to improve the results, it requests more information from the user (e.g., ``Please provide more information.'').
%Which question would be asked depends on the actions taken by the dialogue manager in the interactive SCR system. The dialogue manager is a DQN.
Then the user simulator responds (e.g., ``Trump''). 
The response is determined based on the current search results and the relevant documents.
Given the response, the SCR system updates the search results and poses another question. 
This interaction between the SCR system and the user simulator continues until the simulator decides to terminate (the user is satisfied with the results, or gives up).

During this interaction, the retrieval system does not know which documents are relevant.
It must learn to ask the most efficient question given the situation to obtain the information with least interaction.
For the user simulator, though, although it knows which documents are relevant, it cannot directly access the database.
It can only update the search results by providing useful responses to the SCR system.
The DQN parameters in the dialogue manager and the user simulator are jointly learned so the SCR finds the relevant documents in the most efficient way.
During interaction, real users also do their best to help the SCR system.
Thus we believe that a user simulator learned in this fashion is better than one based on hand-crafted rules.
%=========================================
 
The interactive SCR system is described in Section 2.1, and the user simulator is described in Section 2.2.  
The DQN is used to determine the actions of both the interactive SCR system and the user simulator.
The relevant DQN algorithms are described in Section 2.3. 

\subsection{Interactive SCR System}
The interactive SCR system comprises the retrieval module, the feature extraction module, and the dialogue manager module~\cite{wu2016interactive}.
The user first enters a query \emph{q} into the system. With this user query, and potentially with other extra feedback information from the user during further user-system interaction~\cite{lee2012improved,zhai2001model,tao2006regularized}, the retrieval module generates a list of retrieval results. %在實驗中，記得用一句話講一下 retrieval 是怎麼做的 - Lee
The retrieved results returned from the retrieval module are fed into the feature extraction module, after which they are represented as a feature vector~\cite{wu2016interactive,wen2013interactive}. 
The dialogue manager determines the action based on the extracted features.
There are four possible actions: return documents\footnote{The system says ``Please view the list and select one item relevant to your need''.}, return key term\footnote{The system says ``Is it related to $t^*$?'', where $t^*$ is a key term.}, return request\footnote{The system says ``Please provide more information''.}, and return topic\footnote{The system shows a list of topics and says ``Which topic is related?''.}.
By taking an action and showing the retrieved results based on the current information, the user's response provides the system with more information, enabling the system to update the retrieved results. 
Given the updated results, the dialogue manager 
goes on to the next dialogue round until the user decides to end the interaction.
%The learning strategy of the dialogue manager is DQN, which is a feed-forward network taking the feature vector extracted from retrieval results as input, and evaluate the expected reward of each action. 
%Ultimately, the action with the highest reward would be chosen. 
%the system collects sufficient information to produce high quality retrieval results. In this way, dialogue manager is entitled to determining to show the final result and terminating the dialogue. -> not determine the end any more, right? 

%沒錯！DM 沒有show result了！--Chung

%============== 請看看我寫得對不對 -- Lee ==============
For the system reward, we use a tailored negative reward for each action because no matter what action is taken by the dialogue manager, all feedback actions represent an additional burden to the user. 
To evaluate the retrieval result, we utilize the mean average precision (MAP) as an indicator: %at the current round. 
the better the MAP, the higher reward obtained by the system. %on the contrary, if the MAP decrease, system get a negative reward. 
The return (total reward) $R$ of a dialogue is defined as
\begin{equation}
  R =  \sum_{k=1}^T c_k + \lambda r_k, \label{eq1}
\end{equation}
where $c_k$ is the negative reward from the action taken at the $k$-th turn, and $r_k$ is the MAP improvement after the interaction. 
$\lambda$ is a constant set by the system developer.
%===================================================

\subsection{User Simulator}

%The user simulator is given a set of queries and their corresponding documents.
Given the action of the SCR system, the user simulator offers a response.
In the user simulator, for each action the corresponding decision maker decides on the suitable response. 
The input of the decision maker is a $K$-dimensional feature vector.
The feature vector is the relevance of the top-$K$ documents in the retrieved list from the SCR system, which represents the retrieval quality of the  retrieval results at the current stage. %K 的實際數字實驗再講，但是為什麼是 49 呢？為甚麼不是５０？-- Lee
If the document ranked at the $k$-th position is relevant, the $k$-th dimension is 1; otherwise, the value is zero (the relevance of the returned documents is known only by the user simulator, not the SCR system).

%The decision makers make the decision based on the  $K$ dimensional  vector and a rule-based user simulator. %在實驗中要說這個 simulator 和 Wu 那篇相同
Because the SCR system has four possible actions, there are four decision makers in the user simulator. %and each decision makers have four outputs. The dialogue manager was trained and tested by simulated users. Each simulated user is generated based on a query and its relevant documents provided by the annotators. To compare the performance of different approaches in Sections~\ref{subsec:exp_naive},  \ref{subsec:exp_hand} and \ref{subsec:exp_drl}, the behavior of  the simulated users is set empirically as below.
The four decision makers are described as follows: 
\begin{itemize}
\item\textit{Return Documents}:  
Given a retrieved list, we rank the relevant documents based on their relevance scores from the retrieval system. 
The response is the relevant documents ranked at a specific position.
The decision maker determines the reply by document rankings.  % AMH: 看看這樣寫對不對 
\item\textit{Return Key Term}: If the key term appears in more than 50\% of the relevant documents, the simulated user replies ``YES'' with probabilities of 100\%, 95\%, 90\%, or 85\%. The probability is decided by the decision maker.
\item\textit{Return Request}: Each term $t$ has a score $S(t)=\sum_{d \in \mathcal{R_T}}{N(t,d) ln(1+idf(t))}$, where $N(t,d)$ is the term frequency of term $t$ in the manual transcription of document $d$, $idf(t)$ the inverse document frequency of term $t$ computed from the manual transcriptions of the document collection, and $\mathcal{R_T}$ the real relevant document set provided by the annotators.
All the terms are ranked according to $S(t)$.
% Selecting the term ranked at which position as response is based on decision maker. 
Again, the decision maker determines the reply by term rankings.  % AMH: 看看這樣寫對不對
\item\textit{Return Topic}: 
Given a query, a set of topics is ranked according to their relevance to the query (the relevance is provided by the annotators).
% Selecting the topic ranked at which position is also based on decision maker. 
The decision maker determines the reply by topic rankings.  % AMH: 看看這樣寫對不對
\end{itemize}

If the MAP of the retrieved result is higher than a threshold (thus meeting the user's information need) or the interaction turns exceeds a maximum number (the user gives up), the interaction stops.
If the interaction ends because the MAP is higher than a threshold, the dialogue is deemed a success and the user simulator receives a positive reward; otherwise, the dialogue fails, and a negative reward is given as punishment\footnote{In contrast to the dialogue manager, the user simulator does not receive a negative reward for each action.}.
In this study the threshold is set to 0.6, and the maximum number of turns is 4.

\subsection{Reinforcement Learning} \label{subsec:rl} 

The dialogue manager in the interactive SCR system as well as all the decision makers in the user simulator are deep Q-networks (DQNs)~\cite{mnih2013playing}. %which integrates the advantages of q-learning and neural network. 
The input of each DQN is a feature representation $s$, 
known also as a \textit{state} per reinforcement learning.
The definitions of this feature representation differ between the dialogue manager and decision maker. 
The DQN output dimension equals the number of possible actions $a$ in the action set.
The output corresponding to each action is the state-action value $Q(s, a)$. 
As the possible actions for the DQN in the dialogue manager are \textit{Return Documents}, \textit{Return Key Term}, \textit{Return Request}, and \textit{Return Topic}, the output dimension for the DQN in  dialogue manager is four.
For the user simulator decision makers, the DQN for the Return Key Term decision maker has four outputs (corresponding to 100\%, 95\%, 90\%, and 85\%).
For the remaining three decision makers, each DQN output dimension corresponds to a ranking position (1st, 2nd, 3rd, etc).
All the DQNs (one in the SCR system and four in the user simulator) are trained to maximize the expected return of the system they belong to via the interaction between the interactive SCR system and the user simulator.
To ameliorate the risk of instability when jointly learning the user simulator and retrieval system~\cite{liu2017iterative}, we update the DQNs in the two systems iteratively. %without any pre-training.
That is, we first fix the DQNs in the user simulator, and update the DQN in the SCR system $C$ times.
Then we switch the training agent:
We update the DQNs in the user simulator another $C$ times, while fixing the SCR system.
The above procedure is conducted iteratively.

In addition to the standard DQN algorithm, we use the extended versions.
As the standard DQN algorithm has been observed to overestimate the Q-values~\cite{van2016deep,hasselt2010double}, double DQN is also used in the following experiment, which decouples the selection from the evaluation~\cite{van2016deep,hasselt2010double}.
Dueling DQN~\cite{wang2015dueling} is further applied here. Compared to typical DQN, dueling DQN uses an altered network structure, splitting it into two streams: one learns to provide an estimate of the state value $V(s)$ for every state, and the other calculates the potential advantages 
of each action at a given state.   % AMH: 看看這樣寫對不對
\begin{table*}[]
%\scriptsize
\centering
\caption{MAP and Return for different approaches evaluated on both one-best transcription and lattices}
\label{my-label}
%\resizebox{\textwidth}{!}{%
\begin{tabular}{|c|c|c|l|l|l|l|}
\hline
\multicolumn{3}{|c|}{Approaches} & \multicolumn{2}{c|}{One-best} & \multicolumn{2}{c|}{Lattices} \\ \hline
User simulator & Dialogue manager & \multicolumn{1}{l|}{Input feature} & \multicolumn{1}{c|}{MAP} & \multicolumn{1}{c|}{Return} & \multicolumn{1}{c|}{MAP} & \multicolumn{1}{c|}{Return}  \\ \hline\hline
\multirow{6}{*}{(a) Rule-based} 
 & \multirow{2}{*}{(a-1) DQN } & Raw & 0.5641 & 99.31 & 0.5847 & 112.24 \\ \cline{3-7}  
 &  & Human+Raw & 0.5655 & 107.02 & 0.5790 & 101.99  \\ \cline{2-7} 
 & \multirow{2}{*}{(a-2) Double DQN} & Raw & \textbf{0.5744} & \textbf{107.08} & 0.5911 & 112.14  \\ \cline{3-7} 
 &  & Human+Raw & \textbf{0.5846} & \textbf{115.70} & \textbf{0.5959} & \textbf{102.52}\\ \cline{2-7} 
 & \multirow{2}{*}{(a-3) Dueling DQN} & Raw & 0.5562 & 96.27 & \textbf{0.5860} & \textbf{113.70}  \\ \cline{3-7} 
 &  & Human+Raw & 0.5665 & 105.35 & 0.5784& 96.37 \\ \cline{2-7} 
% & \multirow{2}{*}{(a-4) Dueling double DQN} & Raw & \textbf{0.5756} & \textbf{109.86} & 0.5906 & 109.59    \\ \cline{3-7} 
% &  & Human+Raw & 0.5827 & 112.76 & \textbf{0.5948} &  \textbf{103.87}  \\ 
\hline\hline
\multirow{2}{*}{(b-1) DQN} & \multirow{2}{*}{DQN} & Raw & 0.5758 & 113.02 & 0.6041 & 162.85   \\ \cline{3-7} 
 &  & Human+Raw & 0.5820 & 130.83 & 0.6027 &  162.93  \\ \hline

\multirow{2}{*}{(b-2) Double DQN} & \multirow{2}{*}{Double DQN} & Raw & \textbf{0.6063} & \textbf{190.39} & \textbf{0.6414} & \textbf{224.42} \\ \cline{3-7} 
 &  & Human+Raw & \textbf{0.6066} & \textbf{199.17} & \textbf{0.6375} &  \textbf{222.66}  \\ \hline

\multirow{2}{*}{(b-3) Dueling DQN} & \multirow{2}{*}{Dueling DQN} & Raw & 0.5780 & 108.72 & 0.5864 &  157.92 \\ \cline{3-7} 
 &  & Human+Raw & 0.5757 & 127.49 & 0.5723 &   158.23 \\ 

%\multirow{2}{*}{(b-4) dueling double DQN} & \multirow{2}{*}{dueling double DQN} & Raw & \textbf{0.6042} & \textbf{192.62} & 0.6325 &  216.19   \\ \cline{3-7} 
% &  & Human+Raw & 0.6083 & 189.01 & 0.6316 &   217.55 \\ 
\hline\hline

\multirow{2}{*}{(c-1) Double DQN} & \multirow{2}{*}{Dueling DQN} & Raw & 0.5733 & 134.20 & 0.5678 & 143.83  \\ \cline{3-7} 
 &  &Human+Raw &0.5573 & 111.84 & 0.5905 &  146.83\\ \hline
 
\multirow{2}{*}{(c-2) Dueling DQN} & \multirow{2}{*}{Double DQN} & Raw & \textbf{0.5899} & \textbf{164.00} & \textbf{0.6494} &  \textbf{238.91}   \\ \cline{3-7} 
 &  & Human+Raw & \textbf{0.6139} & \textbf{208.25} & \textbf{0.6420} & \textbf{225.72} \\ \hline
\end{tabular}%
%}
\end{table*}
%我拿掉　 dueling double DQN　，節省一點空間，請自行重標粗體的位置 -- Lee

\section{Experiments}
% 注意：實驗都要用過去式!!!!! -- Lee

\subsection{Data Set}

In the experiments, we used a Mandarin Chinese broadcast news corpus as the spoken document collection from which to retrieve news. The news stories were recorded from radio or TV stations in Taipei from 2001 to 2003, and comprised a total of 5047 news documents covering a total length of 198 hours. For speech recognition, both one-best transcriptions and lattices were generated for spoken content retrieval. Twenty-two graduate students were recruited as the annotators, providing 163 text queries and their relevant spoken documents (not necessarily including the query terms). 

\subsection{Experiment Settings}
Our retrieval module is based on language modeling~\cite{lafferty2001document,chia2010statistical}. After receiving the user feedback in the dialogue, we use the query-regularized mixture model~\cite{lee2012improved,zhai2001model,tao2006regularized}  to generate the new query $q^\prime$. 
The input dimension $K$ of the decision makers was set to 49. %which is identical to the raw relevant feature for dialogue manager. -> This is pointless ... -- Lee
The output dimensions for the four DQNs in the user simulator were all set to four.

All the DQNs had the same hyper-parameters. We used networks with two hidden layers of 1024 nodes. We used relu as the activation function and set the batch size to 256. The initial learning rate was set to 8e-4. MAP was selected as our retrieval evaluation metric. Ten-fold cross validation was performed in all experiments; that is, for each trial, 8 out of 10 query folds were used for training, 1 for parameter tuning (validation set), and the remaining fold for testing.

\begin{figure}[t]
  \centering
  \includegraphics[width=\linewidth]{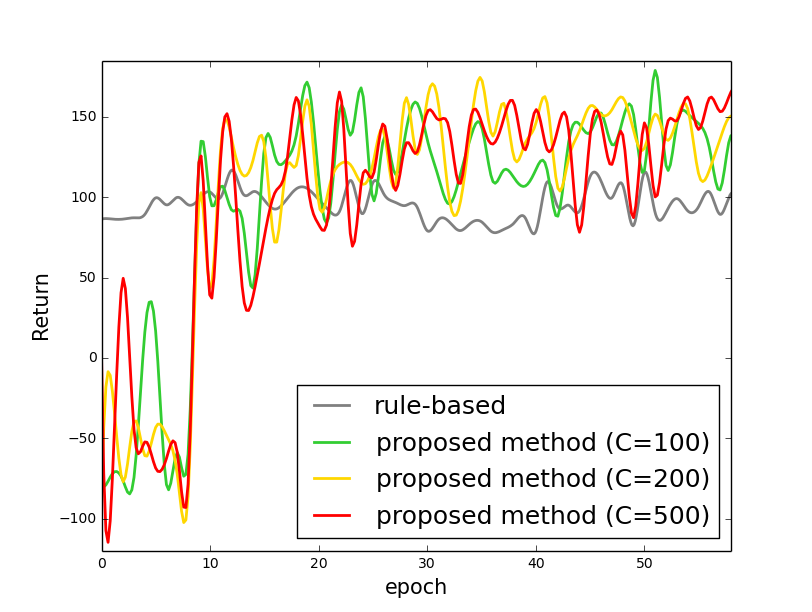}
  \caption{Learning curve of different methods}
  \label{fig:iteration}
\end{figure}

\subsection{Results and Discussion}

In this subsection, the user simulator is used to both train and test the SCR system. 
Figure~\ref{fig:iteration} shows the learning curves of different methods on the training set. 
The rule-based system is identical to previous work~\cite{wu2016interactive}.
The proposed approach is based on a typical DQN 
with the $C$ iterative steps from Section~\ref{subsec:rl}.  % AMH: 看看這樣寫對不對
We find that training with the rule-based system is more stable than the proposed approach, and results in better performance in the first few epochs.
This is reasonable because it is more difficult to train two systems together.
However, with more epochs, the proposed approach eventually outperforms the previous rule-based method.
$C$ was set to 500 in the following experiments.

Table 1 shows the MAP and Return results for the SCR system based on one-best transcription and lattices. %check 這句對不對 -- Lee Ok -Chung
For each  approach, the dialogue manager DQN utilizes two kinds of input features, which were also used in previous work~\cite{wu2016interactive}. 
\textit{Raw} represents the relevance scores of the top-N documents in the retrieved results, and \textit{Human} indicates the use of hand-crafted features based on human knowledge such as clarity score~\cite{he2006query}, ambiguity score~\cite{cronen2002predicting}, and weighted information gain (WIG)~\cite{zhou2007query}. 
The table is divided into three blocks.
In part (a), the rule-based user simulator was used, which is identical to Wu et al.~\cite{wu2016interactive}.
The results of the learnable user simulator are in parts (b) and (c). 
In part (b), we used the same DQN algorithm for the user simulator and dialogue manager, while in part (c), we used different algorithms.

In part (a), we compare the different DQN algorithms with the rule-based user. %which we adopt the previous approach in learning algorithm but use the same reward mechanism as the other experiments to make them comparable. 
The results show that in some cases, both double and dueling DQN outperform the typical DQN.
Using the same DQN algorithm, the trainable user simulator always outperforms the rule-based user simulator (rows (b-1) v.s. (a-1), rows (b-2) v.s. (a-2), rows (b-3) v.s. (a-3)).
For example, compare (b-1) with the baseline (a-1): the trainable user simulator yields improvements of 35.2\% in terms of Return. %check 這句對不對 -- Lee  OK --Chung
%In addition, the synchronous learning using double (b-2) and dueling double DQN (b-4) outperform the baseline by 98.9\% and 94.3\% in average.
We further achieve our best performance on (c-2), by utilizing dueling DQN for the user simulator and double DQN for the dialogue manager. Why this combination yields the best performance is still under investigation.
It shows significant improvement, outperforming the rule-based baseline ((c-2) v.s. (a-1)) by 88.4\% and joint learning using the same DQN algorithm ((b-2) v.s. (a-1)) by 99.0\%.
However, it is not enough to simply show that the trainable user simulator results in higher returns for the SCR system: it is possible that the user simulator is learning a very unnatural way to provide information to the SCR system.
We address this in the next subsection.

%自己 check 上面那些xx% 有沒有算對 -- Lee

\subsection{Human evaluation}

\begin{table}[]
\centering
\caption{KL-divergence between rule-based simulator, learnable simulator and human results, as well as each system's entropy}
\label{KL_entropy}
\begin{tabular}{cccc}
\multirow{2}{*}{KL-divergence} & Rule/human & Ours/human      & Rule/ours \\ \cline{2-4} 
                               & 5.9899     & \textbf{2.1503} & 6.1859    \\
\multirow{2}{*}{Entropy}       & Rule & Ours            & Human     \\ \cline{2-4} 
                               & 0    & \textbf{0.6006} & 0.8886   
\end{tabular}
\end{table}

To verify the correlation and similarity between the proposed user simulator and real users, we performed a subjective human evaluation with 21 subjects, all of whom were graduate students. Given a sampled query, the subject was asked to identify the most relevant document in the top-4 retrieved relevant documents\footnote{Because the output dimension of the decision maker for \textit{Return Document} is 4, it returns only relevant documents ranked in the top four.}. 
Due to space limitations, we show only the results of \textit{Return Document}; the results for the other actions are similar. %是這樣嗎？ -- Lee
%in fact, we cannot find such more instance for topic,request,and keyterm.
%There are 14 multiple choice questions corresponding to 14 sampled query in questionnaire. % 這句看不懂 -- Lee
%這裡想說問卷中的14個題目都是四選一的單選題，14個相對應的query是sample出來的 -> 我還是不太清楚，請直接寫收集到的資料數目 -- Lee
The choices of human, rule-based user, and decision maker are considered random variables (with 1st, 2nd, 3rd, and 4th as the outcomes).
We calculated the KL-divergence between the rule-based user simulator, our proposed trainable user simulator, and the human evaluation result. %  trainable user simulator 是 Table I 的那一個結果？請註明 -- Lee
In Table 2, which lists the average results, the abbreviations are Rule, Ours, and Human respectively. 
The proposed user simulator and human have the smallest KL divergence (Ours/Human).
This shows that the proposed joint training of the user simulator was more closely correlated to the distribution of human evaluations than the rule-based approach. 
We further calculated the entropy of each distribution.
The entropy of the rule-based system was zero because given the same input, it always produces the same response.
The result shows that the decisions made by the proposed user simulator were more diverse and also closer to human evaluation than the rule-based simulator. 
%rule-based 不是只會給一樣的答案嗎？這樣 entropy 不是應該是 0 嗎？怎麼會是0.00034 ？ -- Lee
%因為方便算entropy ，所以把0 用0.00001代替 --Chung 我直接改 0 -- Lee

\section{Conclusions}
User-system interaction is highly desired for SCR. 
Interactive SCR is usually learned using simulated users, but designing simulated users using hand-crafted rules is difficult.
In this paper, we propose a new framework in which the user simulator and the interactive SCR system are jointly learned. 
The experimental results show that the proposed method leads to promising improvements on return. 
Human evaluation further shows that a trainable user simulator much more closely reflects human behavior.

\bibliographystyle{IEEEtran}

\bibliography{mybib}

% \begin{thebibliography}{9}
% \bibitem[1]{Davis80-COP}
%   S.\ B.\ Davis and P.\ Mermelstein,
%   ``Comparison of parametric representation for monosyllabic word recognition in continuously spoken sentences,''
%   \textit{IEEE Transactions on Acoustics, Speech and Signal Processing}, vol.~28, no.~4, pp.~357--366, 1980.
% \bibitem[2]{Rabiner89-ATO}
%   L.\ R.\ Rabiner,
%   ``A tutorial on hidden Markov models and selected applications in speech recognition,''
%   \textit{Proceedings of the IEEE}, vol.~77, no.~2, pp.~257-286, 1989.
% \bibitem[3]{Hastie09-TEO}
%   T.\ Hastie, R.\ Tibshirani, and J.\ Friedman,
%   \textit{The Elements of Statistical Learning -- Data Mining, Inference, and Prediction}.
%   New York: Springer, 2009.
% \bibitem[4]{YourName17-XXX}
%   F.\ Lastname1, F.\ Lastname2, and F.\ Lastname3,
%   ``Title of your INTERSPEECH 2018 publication,''
%   in \textit{Interspeech 2018 -- 19\textsuperscript{th} Annual Conference of the International Speech Communication Association, September 2-6, Hyderabad, India Proceedings, Proceedings}, 2018, pp.~100--104.
% \end{thebibliography}

\end{document}